\documentclass[epjST]{svjour}
\usepackage{color}
\usepackage{amsmath}
\usepackage{amssymb}
\usepackage{graphicx}
\usepackage{subfigure}
\usepackage{bm}

\begin{document}

\title{Digital twins of nonlinear dynamical systems: A perspective}
%\subtitle{Do you have a subtitle?\\ If so, write it here}

\author{Ying-Cheng Lai\inst{1}\fnmsep\thanks{\email{ylai1@asu.edu}} }

\institute{School of Electrical, Computer, and Energy Engineering,
Arizona State University, Tempe, Arizona 85287, USA}

\abstract{
Digital twins have attracted a great deal of recent attention from a wide range of fields. A basic requirement for digital twins of nonlinear dynamical systems is the ability to generate the system evolution and predict potentially catastrophic emergent behaviors so as to providing early warnings. The digital twin can then be used for system ``health'' monitoring in real time and for predictive problem solving. In particular, if the digital twin forecasts a possible system collapse in the future due to parameter drifting as caused by environmental changes or perturbations, an optimal control strategy can be devised and executed as early intervention to prevent the collapse. Two approaches exist for constructing digital twins of nonlinear dynamical systems: sparse optimization and machine learning. The basics of these two approaches are described and their advantages and caveats are discussed. 
} 

\maketitle

\section{Introduction} \label{sec:intro}

In applications it is often the case that an accurate mathematical model of
the underlying dynamical system is not available but time series measurements
or observations of some key variables can be made. If the existing empirical 
data indicate that the underlying system has been functioning as designed or 
``healthy,'' how to anticipate any future potential collapse of the system,
e.g., caused by slow drifting of a system parameter? Digital twins provide a
viable solution. In particular, if a digital ``copy'' of the system can be 
faithfully constructed, then a computational bifurcation analysis with respect
to variations in the parameter of interest can be performed to assess the 
possible future collapse of the system. 

Recent years have witnessed a fast growing interest in building digital twins
not only in  many fields of science and engineering but also in industry,
health care, and defense~\cite{RSK:2020}. Historically, digital twins were 
first used for predicting the structural life of aircraft~\cite{TIES:2011}. 
In dynamical systems, digital twins can be exploited for predicting the future 
states and anticipating emergent, potentially catastrophic 
behaviors~\cite{TQ:2019}. In medicine and health care, for a certain type of 
disease, mechanistic knowledge, observational or diagnostic data, medical 
histories, and detailed physiological modeling can be combined to construct 
patient-specific digital twins~\cite{BSvdH:2018,SWBB:2020,LSG:2021}. 
Development of digital twins of the Earth for green transition is currently
underway in Europe~\cite{Voosen:2020,BSH:2021}.

\begin{figure}
\centering
\resizebox{\columnwidth}{!}{\includegraphics{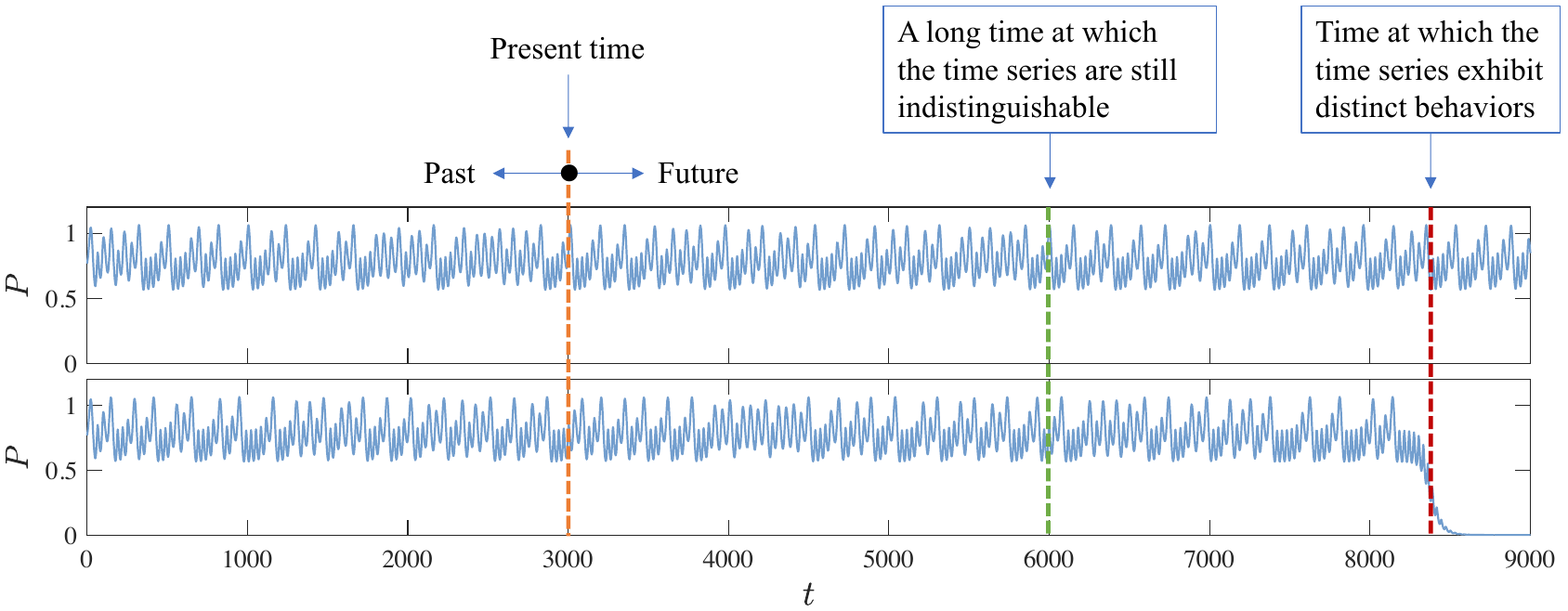}}
\caption{A challenging prediction problem that was previously deemed unsolvable
in nonlinear dynamics. Shown are two time series from a chaotic system at two
different parameter values, respectively. The system exhibits a crisis, a 
global bifurcation that destroys the chaotic attractor, at a critical parameter
value $p_c$. The parameter values corresponding to the time series in the top 
and bottom panels are before and after $p_c$, respectively. In the observation
time interval $[0,3000]$ (corresponding approximately to about 80 oscillation 
cycles of the dynamical variable), the two time series are statistically 
indistinguishable with approximately identical nonzero mean values (not 
extinction). Even when the observation time interval is twice as long
($[0,6000]$), the two time series still cannot be distinguished. Only when 
the observation time extends to over $8000$ (corresponding to about 250 cycles 
of oscillation - the red dashed vertical line) will the time series exhibit 
completely different behavior: one sustained (top) and another collapsed toward 
zero (bottom). Suppose the observation time is $t = 3000$ - the present time, 
so the only information available about the system is the two time series. How 
can the future behaviors of the two time series, i.e., one corresponding to 
sustained or healthy behavior while another to extinction, be predicted based 
on the time series that cannot be distinguished?}   
\label{fig:schematic}       
\end{figure}

The aim of this Perspective is to present an overview of the current approaches
to digital twins for nonlinear dynamical systems. The need for digital 
twins can be appreciated through an illustrative example. As shown in 
Fig.~\ref{fig:schematic}, a dynamical system of interest generates two time
series at two slightly different parameter values: one before a critical 
transition and another after. Before the transition, the system functions 
``normally'' in the sense that the dynamical variable plotted has a finite
mean value, in spite of the statistical fluctuations, as shown in the top 
panel. The variable can be, e.g., the population of a protected species in an 
ecosystem. After the transition, for an initial period of time, the variable
exhibits statistically indistinguishable behaviors from that before the 
transition. However, in the long run the variable becomes zero, signifying, 
e.g., population extinction. If observations were made at any time before
the variable begins to decrease systematically, any observation would suggest 
that the system is completely healthy and functional. Assume that a model of 
the system is not available and all information that can be obtained from the 
system are time series measurements. The question is, if at a time when all 
measurements or observations of the system give no indication of any 
``abnormal'' behavior of the system, how can one tell that in one case the 
system will continue to be functional (the top panel in 
Fig.~\ref{fig:schematic}), but in another case, a catastrophic collapse will 
occur (the bottom panel in Fig.~\ref{fig:schematic}), based on measured time 
series only? This model-free prediction of system's future behavior is an 
extremely challenging problem in applied nonlinear dynamics. Digital twins 
provide a solution.

At the present, there are two main approaches to digital twins in nonlinear
dynamical systems. One is based on reconstructing the system model by finding
the accurate equations governing the dynamical evolution from measurements. 
Crutchfield and McNamara~\cite{CM:1987} pioneered the problem of determining 
the system equations from measurements based on estimating the information 
contained in a sequence of observations to deduce an approximate set of 
equations of motion representing the deterministic portion of the system 
dynamics. Bollt proposed the idea of constructing a dynamical system ``near'' 
the original system with a desired invariant density by exploiting the
Frobenius–Perron theorem~\cite{Bollt:2000}. Later, Yao and Bollt developed
a least-squares approximation strategy to estimate the system model and 
parameters~\cite{YB:2007}. In the past decade or so, a leading approach to
finding system equations~\cite{WYLKG:2011,WLGY:2011,WYLKH:2011,SNWL:2012,SWL:2012,SWL:2014,SLWD:2014,SWFDL:2014,SWWL:2016} 
is based on sparse optimization such as compressive 
sensing~\cite{CRT:2006a,CRT:2006b,Candes:2006,Donoho:2006,Baraniuk:2007,CW:2008}
in situations where these equations have a ``sparse'' structure
~\footnote{The idea of exploiting sparse optimization for discovering system 
equations was first published by the ASU group in 
2011~\cite{WYLKG:2011,WLGY:2011}. Five years later (in 2016), the same idea 
was republished and named as ``SINDy'' [S. L. Brunton, J. L. Proctor, and 
J. Nathan Kutz, ``Discovering governing equations from data by sparse 
identification of nonlinear dynamical systems,'' Proc. Nat. Acad. Sci. 
{\bf 113}, 3932-3937 (2016)]. Approximately five months before this 2016 paper
was published, at a Program Review meeting, Prof. Kutz was made aware of the 
ASU work earlier and was provided the references.}.  
The basic idea is as follows. If the vector fields are smooth, they can be 
approximated by some series expansions such as power or Fourier series. 
The task then becomes that of estimating the various coefficients in the series
expansion. If most of these coefficients are non-zero, the problem is not 
simplified as the total number of coefficients to be determined will be large.
However, if the series expansion is sparse in the sense that the vast majority
of the coefficients are zero, then well-developed sparse-optimization methods
such as compressive sensing can be used to uniquely solve the few non-trivial
coefficients even with a small amount of data~\cite{WYLKG:2011,WLGY:2011}. 
With those coefficients, the system equations described by the series 
expansions represent a ``digital copy'' of the original system.

The second approach to digital twin is machine learning~\cite{KWGHL:2023}.
The basic idea is that a dynamical system functions to evolve the state vector 
forward in time according to a set of mathematical rules, so a digital twin 
must also be able to evolve the state vector forward in time even without any 
input. Reservoir computing~\cite{Jaeger:2001,MNM:2002,JH:2004} is a suitable
choice because its intrinsic recurrent neural network can be trained to execute 
closed-loop, self dynamical evolution with memory. In recent years, there is
a great deal of interest in reservoir computing for predicting chaotic 
systems~\cite{HSRFG:2015,LBMUCJ:2017,PLHGO:2017,LPHGBO:2017,PHGLO:2018,Carroll:2018,NS:2018,ZP:2018,GPG:2019,JL:2019,TYHNKTNNH:2019,FJZWL:2020,ZJQL:2020,KKGGM:2020,KFGL:2021a,PCGPO:2021,KLNPB:2021,FKLW:2021,KFGL:2021b,Bollt:2021,GBGB:2021,Carroll:2022optimizing}. 
The advantage of the machine-learning approach to digital twins is its
applicability to any systems, regardless of the underlying mathematical 
structure of the governing equations (e.g., sparse or dense in terms of some 
series expansion). The disadvantage is that the amount of data required for 
training can be quite demanding.

The sparse-optimization approach to digital twin through discovering system 
equations has been previously reviewed~\cite{WLG:2016,Lai:2021}. The focus of 
this Perspective article is on the general principle of the more recent 
machine-learning approach. 

\section{Digital twins of nonlinear dynamical systems: adaptable machine learning}

Dynamical systems in the real world are not only nonlinear but also complex.
Even if an approximate model of the system can be found, the underlying 
nonlinearity is likely to cause sensitive dependence on initial conditions, 
parameter variations, stochastic fluctuations, and perturbations, rendering 
ineffective any model-based prediction method. To predict characteristic 
changes in the system in advance of their occurrence thus must rely on data 
collected during its normal functioning phase, for which machine learning is 
viable and potentially powerful. 

Most previous studies on reservoir computing focused on the behavior of the
target dynamical system at a fixed parameter setting, i.e., once the machine
has been trained through learning for certain parameter values, it is utilized
to predict the state evolution of the system but at the same set of parameter
values. A digital twin of the system, by its nature, must be able to faithfully
generate the change in the system behavior as some parameter varies. 
A basic requirement of digital twin is that it must be able to generate
the correct bifurcation behaviors of the original system. That is, the digital
twin must not only capture the ``dynamical climate'' of the original system,
but also accurately reflect how the climate changes with the bifurcation or 
control parameter. Adaptable machine learning~\cite{KFGL:2021a,KFGL:2021b}
was developed to meet this challenge, where the term ``adaptable'' was 
introduced to mean that a machine trained with time series data in one 
parameter regime is capable of generating the dynamical behaviors of the 
target system in another, distinct parameter regime. The former is referred
to as the parameter regime of normal system functioning from which the
training data are collected, while the latter is the prediction regime in
which system collapse can occur.

\begin{figure}[ht!]
\centering
\resizebox{\columnwidth}{!}{\includegraphics{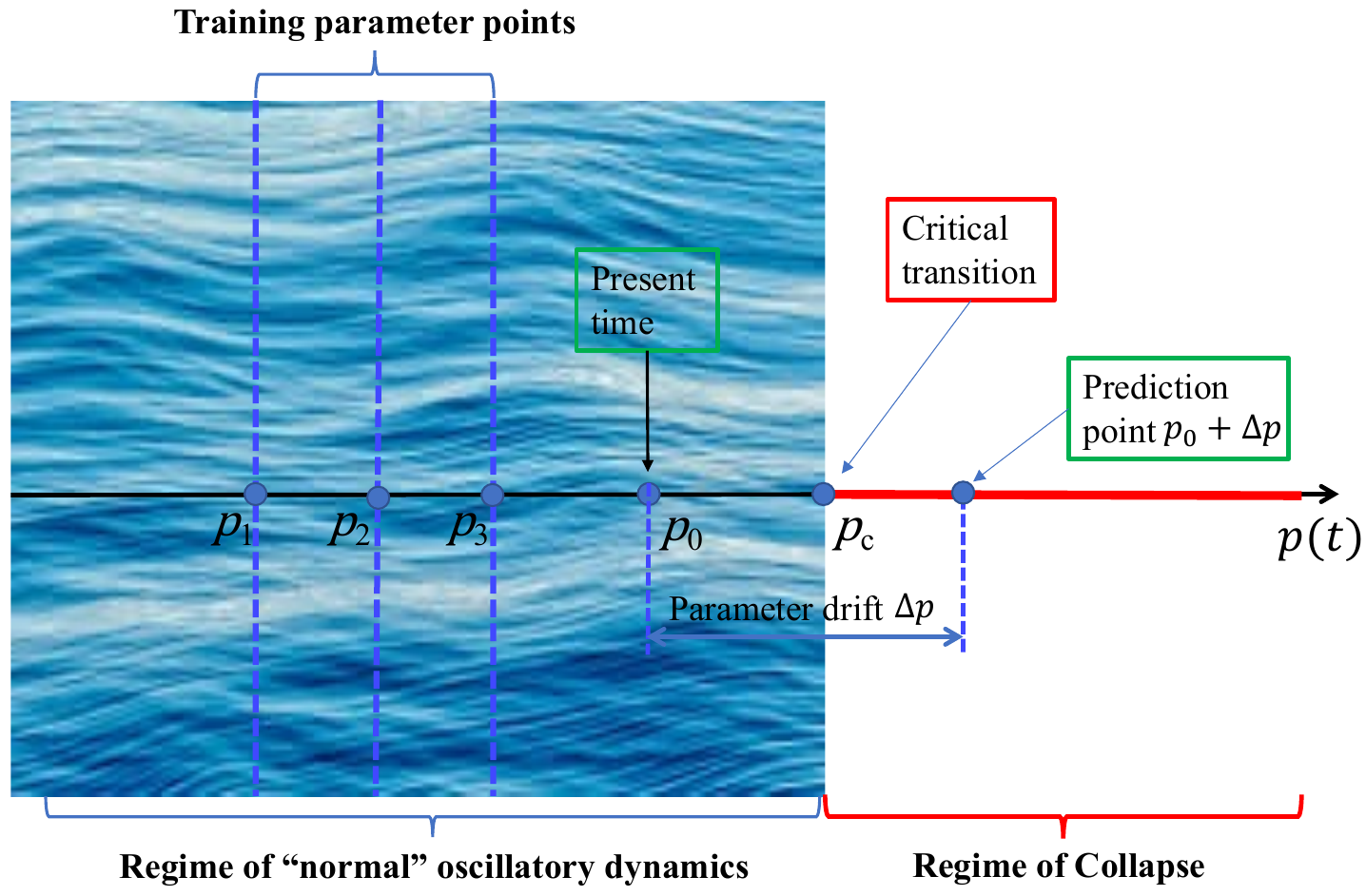}}
\caption{Training scheme of adaptable machine learning.
The target system of interest has two characteristically distinct operational
regimes: normal/oscillatory and collapse regimes which are separated by
a critical transition point $p_c$, where $p$ is a bifurcation parameter. 
As $p$ increases through $p_c$, the system transitions from the normal to
the collapse regime. Suppose the parameter drifts slowly with time, and let
$p_0$ be its value at the present time. The parameter values $p_1$, $p_2$, 
and $p_3$, as indicated by the three vertical blue dashed lines, thus occur 
in the past, from which observational data or time series have been obtained. 
Training of the neural machine is done using these time series in the normal or 
pre-transition regime. The future behavior of the system can be predicted 
by adding a parameter variation $\Delta p$ (corresponding to a specific time
in the future) to $p_0$ and observing the dynamical state of the machine under 
the parameter value $p_0 + \Delta p$. For $p_0 + \Delta p < p_c$, a well
trained machine shall predict that the system will still be in the normal 
functional regime. For $p_0 + \Delta p > p_c$, the machine would generate 
dynamical evolution that is indicative of system collapse.} 
\label{fig:idea}
\end{figure}

The adaptable machine learning framework is schematically shown in 
Fig.~\ref{fig:idea}. Its working principle can be explained, as follows.
Let $p$ be the bifurcation parameter of the target nonlinear system. As $p$ 
varies, a critical point arises: $p_c$, where the system functions normally 
for $p < p_c$ and it exhibits a transient towards collapse for $p > p_c$. 
Training of the digital twin is done based on the time series taken from a 
small number of parameter values in the normal regime, e.g., 
$p_1 < p_2 < p_3 < p_c$. For each parameter value, adequate training is 
required in the sense that the twin is able to predict correctly and 
accurately the oscillatory behavior at the same parameter value for a 
reasonable amount of time. Suppose that, currently, the system functioning 
is normal and it operates at the parameter value $p_0 < p_c$. In the 
prediction phase, suppose a parameter change $\Delta p > 0$ has occurred. The 
new parameter value $p_0 + \Delta p$ is then fed into the digital twin
through the parameter channel. The prediction is deemed successful if the 
twin generates normal oscillations for $p_0 + \Delta p < p_c$ but exhibits 
a transient towards collapse for $p_0 + \Delta p > p_c$.

A recent work demonstrated that the machine-learning architecture of reservoir 
computing is effective as digital twins for a variety of nonlinear dynamical 
systems~\cite{KWGHL:2023}. A reservoir computing machine consists of three main
components: an input layer, a hidden layer with a high-dimensional and complex 
neural network (the reservoir network), and an output layer. The input layer 
maps the typically low-dimensional time series data into the high-dimensional 
state space of the reservoir network, and the output layer projects the 
high-dimensional dynamical evolution of the neural network state back into 
low-dimensional time series (readout). Training is administered to adjust the 
parameters associated with the projection matrix of the output layer to 
minimize the difference between the output and the true input time series. 
Because of the nature of the recurrent neural network, the input matrix
and the reservoir network structure and link weights are chosen {\em a priori} 
according to the values of a few hyperparameters (e.g., the network spectral 
radius) and are fixed during the training and prediction phases. As a result, 
highly efficient learning can be achieved. In terms of hardware realization, 
reservoir computing can be implemented using electronic, time-delay autonomous 
Boolean systems~\cite{HSRFG:2015} or high-speed photonic 
devices~\cite{LBMUCJ:2017}.

\begin{figure}[ht!]
\centering
\resizebox{\columnwidth}{!}{\includegraphics{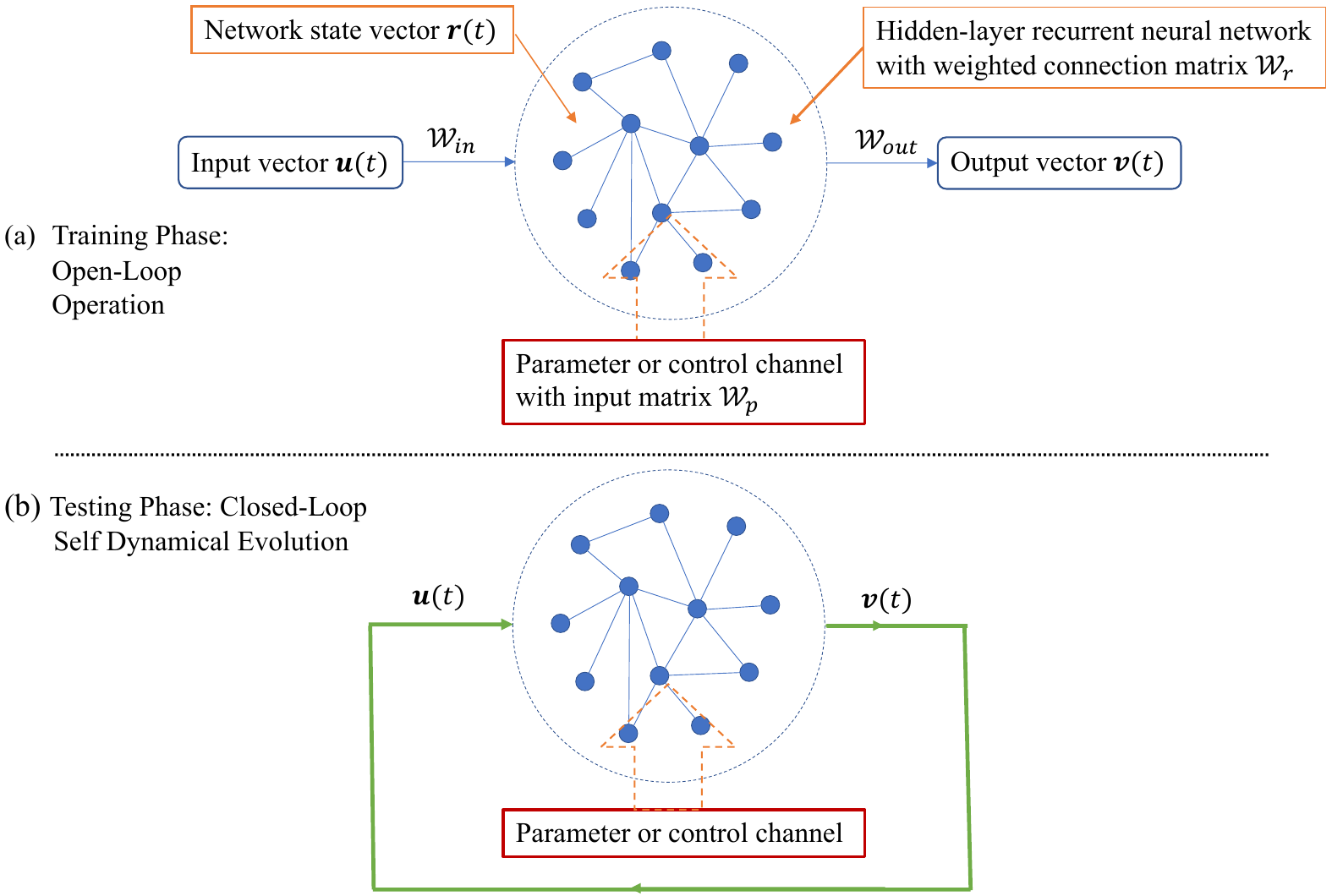}}
\caption{Basic structure of adaptable reservoir computing.
(a) Training phase. Time series data provide the input to the machine.
The input matrix $\mathcal{W}_{in}$ maps the $M$-dimensional input data to a
vector of much higher dimension $N$, where $N \gg M$, and the matrix 
$\mathcal{W}_p$ feeds the bifurcation parameter value into each and every
neuron in the hidden layer as denoted by the dashed circle. The complex neural 
network of $N$ interconnected neurons in the hidden layer is characterized by 
the $N\times N$ weighted matrix $\mathcal{W}_r$. The dynamical state of the 
$i^{th}$ neuron in the reservoir is $r_i$, for $i=1,\ldots,N$, constituting the state vector $\mathbf{r}(t)$. The output matrix $\mathcal{W}_{our}$ converts the
$N$-dimensional state vector of the reservoir network into an $L$-dimensional
output vector, where $N \gg L$. For constructing a digital twin, it is
necessary to set $M=L$. During the training phase, the vector
$\mathbf{u}(t)$ is the input data, so the system is in open-loop operation. 
(b) In the prediction phase, the external input is cut off and the output 
vector $\mathbf{v}(t)$ is directly fed back as the input to the reservoir, 
generating a closed-loop, self-evolving dynamical system.}
\label{fig:RC}
\end{figure}

There are two major types of reservoir computing systems: echo state networks
(ESNs)~\cite{Jaeger:2001} and liquid state machines~\cite{MNM:2002}.
The architecture of an ESN is one that is associated with supervised learning
underlying RNNs. The basic principle of ESNs is to drive a large neural
network of a random or complex topology---the reservoir network---with the
input signal. Each neuron in the network generates a nonlinear response signal.
Linearly combining all the response signals with a set of trainable parameters
yields the output signal. A schematic illustration of the proposed adaptable
reservoir computing scheme is shown in Fig.~\ref{fig:RC}, where the training
and testing configurations are illustrated in Figs.~\ref{fig:RC}(a) and 
\ref{fig:RC}(b), respectively. The machine consists of three components: 
(i) an input layer that maps the low-dimensional ($M$) input signal into a 
(high) $N$-dimensional signal through the weighted $N\times M$ matrix 
$\mathcal{W}_{in}$, (ii) the reservoir network of $N$ neurons characterized by 
$\mathcal{W}_r$, a weighted network matrix of dimension $N\times N$, and 
(iii) an output layer that converts the $N$-dimensional signal from the 
reservoir network into an $L$-dimensional signal through the output weighted 
matrix $\mathcal{W}_{out}$, where $L \sim M \ll N$. The matrix $\mathcal{W}_r$ 
defines the structure of the reservoir neural network in the hidden layer, 
where the dynamics of each node are described by an internal state and a 
nonlinear (e.g., hyperbolic tangent) activation function. For constructing a
digital twin, it is necessary to set $M=L$. As mentioned, the matrices 
$\mathcal{W}_{in}$ and $\mathcal{W}_r$ are generated randomly prior to 
training, whereas all elements of $\mathcal{W}_{out}$ are to be determined 
through training.

Consider the setting where the system and environmental variations are 
characterized by the changes in a single parameter - the ``bifurcation 
parameter.'' The idea is to designate an additional input channel to feed the 
parameter value into each and every artificial neuron in the hidden-layer 
network, as shown in Fig.~\ref{fig:RC}, which makes the reservoir computing 
machine ``cognizant'' of the parameter variations. The basic considerations 
are as follows. To predict critical transitions and system 
collapse, a requirement is that the time series data must be obtained while 
the system is still in normal operation, and it is necessary to collect data 
from multiple values of the bifurcation parameter in the normal phase. Because 
the training data come from several distinct bifurcation parameter values, it 
is necessary that the machine ``know'' the parameter values at which the data 
are taken, which can be accomplished by ``injecting'' the parameter value to 
all nodes of the recurrent dynamical neural network in the hidden layer.

\section{Examples of digital twins of nonlinear dynamical systems} \label{sec:DT_examples}

\subsection{Systems for which sparse optimization methods fail}

Recall that the basic requirement of any sparse optimization technique for
finding the system equations is {\em sparsity}: when the system equations are
expanded into a power series or a Fourier series, it must be that only a few
terms are present so that the coefficient vectors to be determined from data
are sparse~\cite{WYLKG:2011,WLG:2016}. However, there are physical and 
biological systems that violate this sparsity requirement. An example is the  
two-dimensional Ikeda map describing the dynamics of a laser pulse propagating 
in a nonlinear cavity~\cite{Ikeda:1979,IDA:1980,HJM:1985}:
\begin{eqnarray} \label{eq:Ikeda}
z_{n+1}= \mu+\gamma z_n \exp{\left( i\kappa-\frac{i\nu}{1+|z_n|^2} \right)},
\end{eqnarray}
where the dynamical variables $x$ and $y$ are the real and imaginary parts
of the complex variable $z$, $\mu$ is the dimensionless laser input amplitude
(a convenient bifurcation parameter), $\gamma$ is the reflection coefficient 
of the partially reflecting mirrors of the cavity, $\kappa$ is the cavity 
detuning parameter, and $\nu$ characterizes the detuning contributed by the 
nonlinear medium in the cavity. If the map functions are expanded into a 
power series or a Fourier series, an infinite number of terms will be present.
In fact, for the Ikeda map it remains infeasible to find a suitable 
mathematical base to expand the map functions into a sparse series, rendering 
inapplicable the sparse optimization method for constructing a digital twin. It
was demonstrated~\cite{KFGL:2021b} that adaptable reservoir computing provides 
an effective approach to creating a digital twin of the Ikeda map, which can 
be used to predict bifurcation behaviors and critical transitions of the 
optical-cavity system.

Another example is a three-species ecosystem described by~\cite{MY:1994}
\begin{align} 
\nonumber
\frac{dR}{dt} & =  R(1-\frac{R}{K})-\frac{x_cy_cCR}{R+R_0}, \\ \label{eq:food_chain}
\frac{dC}{dt} & =  x_cC[\frac{y_cR}{R+R_0}-1]-\frac{x_py_pPC}{C+C_0}, \\ \nonumber
\frac{dP}{dt} & =  x_pP(\frac{y_pC}{C+C_0}-1),
\end{align}
where the dynamical variables $R$, $C$, $P$ are the population densities of 
the three species: resource, consumer, and predator, respectively, and the
system parameters are $K$ (the carrying capacity), $x_c$, $y_c$, $x_p$, $y_p$, 
$R_0$, and $C_0$. For a wide range of the parameter values, the system exhibits
a critical transition to species extinction. A power-series expansion of the 
vector field on the right side of Eq.~\eqref{eq:food_chain} contains an infinite
number terms, rendering inapplicable any sparse optimization method. It was 
demonstrated~\cite{KFGL:2021a} that adaptable reservoir computing can be used
to construct a digital twin of the ecosystem to predict the critical transition
and the dynamical behaviors about the transition.

\subsection{Predicting amplitude death}

In nonlinear dynamical systems, it can happen that, when a bifurcation 
parameter of the system changes through a critical point, the oscillatory 
behaviors of the state variables halt suddenly and completely - a phenomenon
called amplitude death~\cite{SPR:2012,KVK:2013a}. From the point of view of 
bifurcation, amplitude death is caused by a sudden transition of the system 
from an oscillatory state to a steady state. If the normal function of
the system relies on oscillations, then this phenomenon will be undesired
and it is important to be able to predict amplitude death before its actual
occurrence. For example, in biological systems, normal conditions are often
associated with oscillations, and amplitude death marks the onset of 
pathological conditions. To anticipate amplitude death in advance of its 
occurrence based on oscillatory time series collected during normal functioning
is important. It was demonstrated that adaptable reservoir computing as a 
digital twin of the system of interest can be effective for this prediction 
task~\cite{XKSL:2021}.  

\subsection{Predicting onset of synchronization}

In complex dynamical systems consisting of a number of coupling elements, 
synchronization is coherent motion among the elements. Depending on the 
specific form of the coherent motion, different types of synchronization can 
emerge, including complete chaotic synchronization~\cite{PC:1990}, phase 
synchronization~\cite{RPK:1996}, and generalized synchronization~\cite{KP:1996}.
The occurrence of synchronization has significant consequences for the system 
behavior and functions. An example is the occurrence of epileptic seizures in 
the brain neural system, where a widely adopted assumption is that 
hypersynchrony is closely associated with the occurrence of epileptic 
seizures~\cite{KSJ:book}, during which the number of independent degrees of 
freedom of the underlying brain dynamical system is reduced. In the extensive 
literature in this field, there was demonstration that partial and transient 
phase synchrony can be exploited to detect and characterize (but not to 
predict) seizure from multichannel brain data~\cite{LFO:2006,LFOH:2007,OL:2011}.
Reliable seizure prediction remains a challenge. In general, it is of interest 
to predict or anticipate synchronization before its actual occurrence based on 
time series data obtained before the system evolves into some kind of 
synchronous dynamical state. In particular, given that the system operates 
in a parameter regime where there is no synchronization, would it be possible 
to predict, without relying on any model, the onset of synchronization based 
solely on the dynamically incoherent time series measurements taken from the 
parameter regime of desynchronization? A digital twin of the original system
represents a viable solution.

It was demonstrated~\cite{FKLW:2021} that adaptable reservoir computing can
be used to construct a digital twin for predicting synchronization. In 
particular, the digital twin can predict, with a given amount of parameter 
change, whether the system would remain asynchronous or exhibit synchronous 
dynamics. Systems tested include representative chaotic and network systems 
that exhibit continuous (second-order) or abrupt (first-order) transitions. Of 
special interest are network dynamical systems exhibiting an explosive 
(first-order) transition and a hysteresis loop, and it was 
shown~\cite{FKLW:2021} that the digital twin possesses the power to
accurately predict these features including the precise locations of the 
transition points associated with the forward and backward transition paths. 

\section{Discussion and outlook} \label{sec:discussion}

There exist two approaches to digital twins in nonlinear dynamical systems: 
sparse optimization and machine learning, where the former relies on finding
the exact governing equations of the system and its applicability is thus 
limited. This Perspective explains the difficulty with the sparse-optimization
approach and focuses on the machine-learning approach. An issue concerns 
the type of machine-learning scheme that can be exploited for constructing 
digital twins for nonlinear dynamical systems. Since a dynamical system  
evolves its state forward in time according to a set of mathematical rules,
its digital twin must be able to evolve forward in time by itself. In this
regard, reservoir computing is capable of closed-loop, self dynamical evolution
with memory, so it provides a base for developing digital twins of nonlinear 
dynamical systems.

An important contribution to explainable machine learning as applied to 
nonlinear dynamical system is the mathematical understanding of the inner
workings of reservoir computing by Bollt~\cite{Bollt:2021}, leading to the 
development of ``next-generation reservoir computing''~\cite{GBGB:2021}.
A foundational problem underlying the development of a physical understanding
of the workings of reservoir-computing based digital twin is searching for 
scaling laws between the complexities of a chaotic system and its digital 
twin. In particular, in order for the digital twin to predict the state 
evolution of the target system, the complexity of the former must ``overpower''
that of the latter. What is the meaning of ``overpowering'' and how can it be 
characterized? Are there scaling laws quantifying the relationship? Answers 
to these questions will provide a deeper understanding of the inner workings 
of reservoir-computing based digital twin.

For a chaotic system, its state evolution is determined by the trajectory
movement on a dynamically invariant set, e.g., a chaotic attractor. The
complexity of the chaotic system can be faithfully characterized by the
information dimension of the chaotic invariant set~\cite{Ott:book,LT:book}.
Likewise, the complexity of the digital twin is determined by its ``inner'' 
dynamical system, which is typically a complex dynamical network in the 
hidden layer of the reservoir computer. For a complex network, in general its 
complexity increases with its size. As the information dimension of the target 
chaotic system increases, the size of the reservoir network must increase
accordingly to warrant its predictive power over the former. A universal
scaling law between the network size required for accurate prediction and
the information dimension of the chaotic system, if it indeed exists, would
represent a meaningful way to characterize the digital twin's overpowering 
the target chaotic system. 

\section*{Data Availability Statement}

No Data associated in the manuscript.

\section*{Acknowledgment}

I thank L.-W. Kong for discussions and for assisting with Fig.~1. This work 
was supported by the Army Research Office through Grant No.~W911NF-21-2-0055.

%\bibliographystyle{ScienceAdvances}
%\bibliography{DT}

\begin{thebibliography}{10}

\bibitem{RSK:2020}
A.~Rasheed, O.~San, T.~Kvamsdal, Digital twin: Values, challenges and enablers
  from a modeling perspective.
\newblock {\it IEEE Access\/} {\bf 8}, 21980-22012 (2020).

\bibitem{TIES:2011}
E.~J. Eric J.~Tuegel, A.~R. Ingraffea, T.~G. Eason, S.~M. Spottswood,
  Reengineering aircraft structural life prediction using a digital twin.
\newblock {\it Int. J. Aerospace Eng.\/} {\bf 2011}, 154798 (2011).

\bibitem{TQ:2019}
F.~Tao, Q.~Qi, Make more digital twins.
\newblock {\it Nature\/} {\bf 573}, 274-277 (2019).

\bibitem{BSvdH:2018}
K.~Bruynseels, F.~S. de~Sio, J.~van~den Hoven, Digital twins in health care:
  {Ethical} implications of an emerging engineering paradigm.
\newblock {\it Front. Gene.\/} {\bf 9}, 31 (2018).

\bibitem{SWBB:2020}
S.~M. Schwartz, K.~Wildenhaus, A.~Bucher, B.~Byrd, Digital twins and the
  emerging science of self: Implications for digital health experience design
  and ``small'' data.
\newblock {\it Front. Comp. Sci.\/} {\bf 2}, 31 (2020).

\bibitem{LSG:2021}
R.~Laubenbacher, J.~P. Sluka, J.~A. Glazier, Using digital twins in viral
  infection.
\newblock {\it Science\/} {\bf 371}, 1105-1106 (2021).

\bibitem{Voosen:2020}
P.~Voosen, Europe builds ‘digital twin’ of earth to hone climate forecasts.
\newblock {\it Science\/} {\bf 370}, 16-17 (2020).

\bibitem{BSH:2021}
P.~Bauer, B.~Stevens, W.~Hazeleger, A digital twin of earth for the green
  transition.
\newblock {\it Nat. Clim. Change\/} {\bf 11}, 80-83 (2021).

\bibitem{CM:1987}
J.~P. Crutchfield, B.~McNamara, Equations of motion from a data series.
\newblock {\it Complex Sys.\/} {\bf 1}, 417-452 (1987).

\bibitem{Bollt:2000}
E.~M. Bollt, Controlling chaos and the inverse frobenius-perron problem: global
  stabilization of arbitrary invariant measures.
\newblock {\it Int. J. Bif. Chaos\/} {\bf 10}, 1033-1050 (2000).

\bibitem{YB:2007}
C.~Yao, E.~M. Bollt, Modeling and nonlinear parameter estimation with
  {Kronecker} product representation for coupled oscillators and spatiotemporal
  systems.
\newblock {\it Physica D\/} {\bf 227}, 78-99 (2007).

\bibitem{WYLKG:2011}
W.-X. Wang, R.~Yang, Y.-C. Lai, V.~Kovanis, C.~Grebogi, Predicting catastrophes
  in nonlinear dynamical systems by compressive sensing.
\newblock {\it Phys. Rev. Lett.\/} {\bf 106}, 154101 (2011).

\bibitem{WLGY:2011}
W.-X. Wang, Y.-C. Lai, C.~Grebogi, J.-P. Ye, Network reconstruction based on
  evolutionary-game data via compressive sensing.
\newblock {\it Phys. Rev. X\/} {\bf 1}, 021021 (2011).

\bibitem{WYLKH:2011}
W.-X. Wang, R.~Yang, Y.-C. Lai, V.~Kovanis, M.~A.~F. Harrison,
  Time-series-based prediction of complex oscillator networks via compressive
  sensing.
\newblock {\it EPL (Europhys. Lett.)\/} {\bf 94}, 48006 (2011).

\bibitem{SNWL:2012}
R.-Q. Su, X.~Ni, W.-X. Wang, Y.-C. Lai, Forecasting synchronizability of
  complex networks from data.
\newblock {\it Phys. Rev. E\/} {\bf 85}, 056220 (2012).

\bibitem{SWL:2012}
R.-Q. Su, W.-X. Wang, Y.-C. Lai, Detecting hidden nodes in complex networks
  from time series.
\newblock {\it Phys. Rev. E\/} {\bf 85}, 065201 (2012).

\bibitem{SWL:2014}
R.-Q. Su, Y.-C. Lai, X.~Wang, Identifying chaotic fitzhugh-nagumo neurons using
  compressive sensing.
\newblock {\it Entropy\/} {\bf 16}, 3889-3902 (2014).

\bibitem{SLWD:2014}
R.-Q. Su, Y.-C. Lai, X.~Wang, Y.-H. Do, Uncovering hidden nodes in complex
  networks in the presence of noise.
\newblock {\it Sci. Rep.\/} {\bf 4}, 3944 (2014).

\bibitem{SWFDL:2014}
Z.~Shen, W.-X. Wang, Y.~Fan, Z.~Di, Y.-C. Lai, Reconstructing propagation
  networks with natural diversity and identifying hidden sources.
\newblock {\it Nat. Commun.\/} {\bf 5}, 4323 (2014).

\bibitem{SWWL:2016}
R.-Q. Su, W.-W. Wang, X.~Wang, Y.-C. Lai, Data based reconstruction of complex
  geospatial networks, nodal positioning, and detection of hidden node.
\newblock {\it R. Soc. Open Sci.\/} {\bf 3}, 150577 (2016).

\bibitem{CRT:2006a}
E.~Cand\`{e}s, J.~Romberg, T.~Tao, Robust uncertainty principles: exact signal
  reconstruction from highly incomplete frequency information.
\newblock {\it IEEE Trans. Info. Theory\/} {\bf 52}, 489-509 (2006).

\bibitem{CRT:2006b}
E.~Cand\`{e}s, J.~Romberg, T.~Tao, Stable signal recovery from incomplete and
  inaccurate measurements.
\newblock {\it Comm. Pure Appl. Math.\/} {\bf 59}, 1207-1223 (2006).

\bibitem{Candes:2006}
E.~Cande\`s, {\it Proceedings of the International Congress of
  Mathematicians\/} (Madrid, Spain, 2006), vol.~3, pp. 1433--1452.

\bibitem{Donoho:2006}
D.~Donoho, Compressed sensing.
\newblock {\it IEEE Trans. Info. Theory\/} {\bf 52}, 1289-1306 (2006).

\bibitem{Baraniuk:2007}
R.~G. Baraniuk, Compressed sensing.
\newblock {\it IEEE Signal Process. Mag.\/} {\bf 24}, 118-121 (2007).

\bibitem{CW:2008}
E.~Cande\`s, M.~Wakin, An introduction to compressive sampling.
\newblock {\it IEEE Signal Process. Mag.\/} {\bf 25}, 21-30 (2008).

\bibitem{KWGHL:2023}
L.-W. Kong, Y.~Weng, B.~Glaz, M.~Haile, Y.-C. Lai, Reservoir computing as
  digital twins for nonlinear dynamical systems.
\newblock {\it Chaos\/} {\bf 33}, 033111 (2023).

\bibitem{Jaeger:2001}
H.~Jaeger, The “echo state” approach to analysing and training recurrent
  neural networks-with an erratum note.
\newblock {\it German National Research Center for Information Technology GMD
  Technical Report\/} {\bf 148}, 13 (2001).

\bibitem{MNM:2002}
W.~Mass, T.~Nachtschlaeger, H.~Markram, Real-time computing without stable
  states: A new framework for neural computation based on perturbations.
\newblock {\it Neur. Comp.\/} {\bf 14}, 2531-2560 (2002).

\bibitem{JH:2004}
H.~Jaeger, H.~Haas, Harnessing nonlinearity: Predicting chaotic systems and
  saving energy in wireless communication.
\newblock {\it Science\/} {\bf 304}, 78-80 (2004).

\bibitem{HSRFG:2015}
N.~D. Haynes, M.~C. Soriano, D.~P. Rosin, I.~Fischer, D.~J. Gauthier, Reservoir
  computing with a single time-delay autonomous {Boolean} node.
\newblock {\it Phys. Rev. E\/} {\bf 91}, 020801 (2015).

\bibitem{LBMUCJ:2017}
L.~Larger, {\it et~al.\/}, High-speed photonic reservoir computing using a
  time-delay-based architecture: Million words per second classification.
\newblock {\it Phys. Rev. X\/} {\bf 7}, 011015 (2017).

\bibitem{PLHGO:2017}
J.~Pathak, Z.~Lu, B.~Hunt, M.~Girvan, E.~Ott, Using machine learning to
  replicate chaotic attractors and calculate {Lyapunov} exponents from data.
\newblock {\it Chaos\/} {\bf 27}, 121102 (2017).

\bibitem{LPHGBO:2017}
Z.~Lu, {\it et~al.\/}, Reservoir observers: Model-free inference of unmeasured
  variables in chaotic systems.
\newblock {\it Chaos\/} {\bf 27}, 041102 (2017).

\bibitem{PHGLO:2018}
J.~Pathak, B.~Hunt, M.~Girvan, Z.~Lu, E.~Ott, Model-free prediction of large
  spatiotemporally chaotic systems from data: A reservoir computing approach.
\newblock {\it Phys. Rev. Lett.\/} {\bf 120}, 024102 (2018).

\bibitem{Carroll:2018}
T.~L. Carroll, Using reservoir computers to distinguish chaotic signals.
\newblock {\it Phys. Rev. E\/} {\bf 98}, 052209 (2018).

\bibitem{NS:2018}
K.~Nakai, Y.~Saiki, Machine-learning inference of fluid variables from data
  using reservoir computing.
\newblock {\it Phys. Rev. E\/} {\bf 98}, 023111 (2018).

\bibitem{ZP:2018}
Z.~S. Roland, U.~Parlitz, Observing spatio-temporal dynamics of excitable media
  using reservoir computing.
\newblock {\it Chaos\/} {\bf 28}, 043118 (2018).

\bibitem{GPG:2019}
A.~Griffith, A.~Pomerance, D.~J. Gauthier, Forecasting chaotic systems with
  very low connectivity reservoir computers.
\newblock {\it Chaos\/} {\bf 29}, 123108 (2019).

\bibitem{JL:2019}
J.~Jiang, Y.-C. Lai, Model-free prediction of spatiotemporal dynamical systems
  with recurrent neural networks: Role of network spectral radius.
\newblock {\it Phys. Rev. Research\/} {\bf 1}, 033056 (2019).

\bibitem{TYHNKTNNH:2019}
G.~Tanaka, {\it et~al.\/}, Recent advances in physical reservoir computing: A
  review.
\newblock {\it Neu. Net.\/} {\bf 115}, 100--123 (2019).

\bibitem{FJZWL:2020}
H.~Fan, J.~Jiang, C.~Zhang, X.~Wang, Y.-C. Lai, Long-term prediction of chaotic
  systems with machine learning.
\newblock {\it Phys. Rev. Research\/} {\bf 2}, 012080 (2020).

\bibitem{ZJQL:2020}
C.~Zhang, J.~Jiang, S.-X. Qu, Y.-C. Lai, Predicting phase and sensing phase
  coherence in chaotic systems with machine learning.
\newblock {\it Chaos\/} {\bf 30}, 083114 (2020).

\bibitem{KKGGM:2020}
C.~Klos, Y.~F.~K. Kossio, S.~Goedeke, A.~Gilra, R.-M. Memmesheimer, Dynamical
  learning of dynamics.
\newblock {\it Phys. Rev. Lett.\/} {\bf 125}, 088103 (2020).

\bibitem{KFGL:2021a}
L.-W. Kong, H.-W. Fan, C.~Grebogi, Y.-C. Lai, Machine learning prediction of
  critical transition and system collapse.
\newblock {\it Phys. Rev. Research\/} {\bf 3}, 013090 (2021).

\bibitem{PCGPO:2021}
D.~Patel, D.~Canaday, M.~Girvan, A.~Pomerance, E.~Ott, Using machine learning
  to predict statistical properties of non-stationary dynamical processes:
  System climate, regime transitions, and the effect of stochasticity.
\newblock {\it Chaos\/} {\bf 31}, 033149 (2021).

\bibitem{KLNPB:2021}
J.~Z. Kim, Z.~Lu, E.~Nozari, G.~J. Pappas, D.~S. Bassett, Teaching recurrent
  neural networks to infer global temporal structure from local examples.
\newblock {\it Nat. Machine Intell.\/} {\bf 3}, 316--323 (2021).

\bibitem{FKLW:2021}
H.~Fan, L.-W. Kong, Y.-C. Lai, X.~Wang, Anticipating synchronization with
  machine learning.
\newblock {\it Phys. Rev. Resesearch\/} {\bf 3}, 023237 (2021).

\bibitem{KFGL:2021b}
L.-W. Kong, H.~Fan, C.~Grebogi, Y.-C. Lai, Emergence of transient chaos and
  intermittency in machine learning.
\newblock {\it J. Phys. Complexity\/} {\bf 2}, 035014 (2021).

\bibitem{Bollt:2021}
E.~Bollt, On explaining the surprising success of reservoir computing
  forecaster of chaos? the universal machine learning dynamical system with
  contrast to var and dmd.
\newblock {\it Chaos\/} {\bf 31}, 013108 (2021).

\bibitem{GBGB:2021}
D.~J. Gauthier, E.~Bollt, A.~Griffith, W.~A. Barbosa, Next generation reservoir
  computing.
\newblock {\it Nat. Commun.\/} {\bf 12}, 1--8 (2021).

\bibitem{Carroll:2022optimizing}
T.~L. Carroll, Optimizing memory in reservoir computers.
\newblock {\it Chaos\/} {\bf 32}, 023123 (2022).

\bibitem{WLG:2016}
W.-X. Wang, Y.-C. Lai, C.~Grebogi, Data based identification and prediction of
  nonlinear and complex dynamical systems.
\newblock {\it Phys. Rep.\/} {\bf 644}, 1-76 (2016).

\bibitem{Lai:2021}
Y.-C. Lai, Finding nonlinear system equations and complex network structures
  from data: {A} sparse optimization approach.
\newblock {\it Chaos\/} {\bf 31}, 082101 (2021).

\bibitem{Ikeda:1979}
K.~Ikeda, Multiple-valued stationary state and its instability of the
  transmitted light by a ring cavity system.
\newblock {\it Opt. Commun.\/} {\bf 30}, 257-261 (1979).

\bibitem{IDA:1980}
K.~Ikeda, H.~Daido, O.~Akimoto, Optical turbulence: Chaotic behavior of
  transmitted light from a ring cavity.
\newblock {\it Phys. Rev. Lett.\/} {\bf 45}, 709--712 (1980).

\bibitem{HJM:1985}
S.~M. Hammel, C.~K. R.~T. Jones, J.~V. Moloney, Global dynamical behavior of
  the optical field in a ring cavity.
\newblock {\it J. Opt. Soc. Ame. B\/} {\bf 2}, 552--564 (1985).

\bibitem{MY:1994}
K.~McCann, P.~Yodzis, Nonlinear dynamics and population disappearances.
\newblock {\it Ame. Naturalist\/} {\bf 144}, 873--879 (1994).

\bibitem{SPR:2012}
G.~Saxena, A.~Prasad, R.~Ramaswamy, Amplitude death: The emergence of
  stationarity in coupled nonlinear systems.
\newblock {\it Phys. Rep.\/} {\bf 521}, 205--228 (2012).

\bibitem{KVK:2013a}
A.~Koseska, E.~Volkov, J.~Kurths, Oscillation quenching mechanisms: Amplitude
  vs. oscillation death.
\newblock {\it Phys. Rep.\/} {\bf 531}, 173--199 (2013).

\bibitem{XKSL:2021}
R.~Xiao, L.-W. Kong, Z.-K. Sun, Y.-C. Lai, Predicting amplitude death with
  machine learning.
\newblock {\it Phys. Rev. E\/} {\bf 104}, 014205 (2021).

\bibitem{PC:1990}
L.~M. Pecora, T.~L. Carroll, Synchronization in chaotic systems.
\newblock {\it Phys. Rev. Lett.\/} {\bf 64}, 821--824 (1990).

\bibitem{RPK:1996}
M.~G. Rosenblum, A.~S. Pikovsky, J.~Kurths, Phase synchronization of chaotic
  oscillators.
\newblock {\it Phys. Rev. Lett.\/} {\bf 76}, 1804--1807 (1996).

\bibitem{KP:1996}
L.~Kocarev, U.~Parlitz, Generalized synchronization, predictability, and
  equivalence of unidirectionally coupled dynamical systems.
\newblock {\it Phys. Rev. Lett.\/} {\bf 76}, 1816--1819 (1996).

\bibitem{KSJ:book}
E.~R. Kandel, J.~H. Schwartz, T.~M. Jessell, {\it {Principle of Neural
  Science}\/} (Appleton and Lange, Norwalk CT, 1991), third edn.

\bibitem{LFO:2006}
Y.-C. Lai, M.~G. Frei, I.~Osorio, Detecting and characterizing phase
  synchronization in nonstationary dynamical systems.
\newblock {\it Phys. Rev. E\/} {\bf 73}, 026214 (2006).

\bibitem{LFOH:2007}
Y.-C. Lai, M.~G. Frei, I.~Osorio, L.~Huang, Characterization of synchrony with
  applications to epileptic brain signals.
\newblock {\it Phys. Rev. Lett.\/} {\bf 98}, 108102 (2007).

\bibitem{OL:2011}
I.~Osorio, Y.-C. Lai, A phase-synchronization and random-matrix based approach
  to multichannel time-series analysis with application to epilepsy.
\newblock {\it Chaos\/} {\bf 21}, 033108 (2011).

\bibitem{Ott:book}
E.~Ott, {\it Chaos in Dynamical Systems\/} (Cambridge University Press,
  Cambridge, UK, 2002), second edn.

\bibitem{LT:book}
Y.-C. Lai, T.~T\'{e}l, {\it Transient Chaos - Complex Dynamics on Finite Time
  Scales\/} (Springer, New York, 2011).

\end{thebibliography}

\end{document}